\documentclass[runningheads]{llncs}

\usepackage{eccv}

\usepackage{eccvabbrv}

\usepackage{graphicx}
\usepackage{booktabs}
\usepackage{xcolor}

\usepackage[accsupp]{axessibility}  

\usepackage{hyperref}

\usepackage{orcidlink}

\def\ourmethod{\textsc{AV-MSF}\xspace}

\begin{document}
\title{Objects as Audio-Visual Modal Sound Fields}

\titlerunning{Objects as Audio-Visual Modal Sound Fields}

\author{
Zisen Shao$^*$ \and
Zihao Wei$^*$ \and
Derong Jin \and
Ruohan Gao
}

\authorrunning{Z.~Shao et al.}

\institute{
University of Maryland, College Park\\
\email{\{zisen, zihaowei, djin77, rhgao\}@umd.edu}
\textcolor{magenta}{\url{https://zisenshao.github.io/AV-MSF/}}
}

\maketitle
\begingroup
\renewcommand{\thefootnote}{*}
\footnotetext{Equal contribution.}
\endgroup

\begin{abstract}
    While modern 3D reconstruction excels at modeling object geometry and appearance, it largely ignores the rich acoustic cues revealed through physical interaction. Object impact sounds convey material, stiffness, and structural properties that complement vision, yet existing impact sound modeling approaches either rely on expensive physics-based simulation or require large datasets to generalize in a purely data-driven manner. 
    We introduce Audio-Visual Modal Sound Field (\ourmethod), a novel object-level acoustic representation reconstructed from multi-view images and only a few impact sound recordings. \ourmethod builds on 3D Gaussian Splatting integrated with dense 3D visual feature to provide a strong geometry-aware prior, and represents the impact sound field using compact, physically meaningful modal parameters, enabling robust few-shot reconstruction. 
    Experiments on two real-world datasets show that \ourmethod achieves state-of-the-art impact sound rendering, outperforming both physics-based and data-driven baselines. Furthermore, we demonstrate downstream applications enabled by our representation, including contact localization and object sound editing.
\end{abstract}
\section{Introduction}

\begin{figure}[tb]
  \centering
  \includegraphics[width=\linewidth]{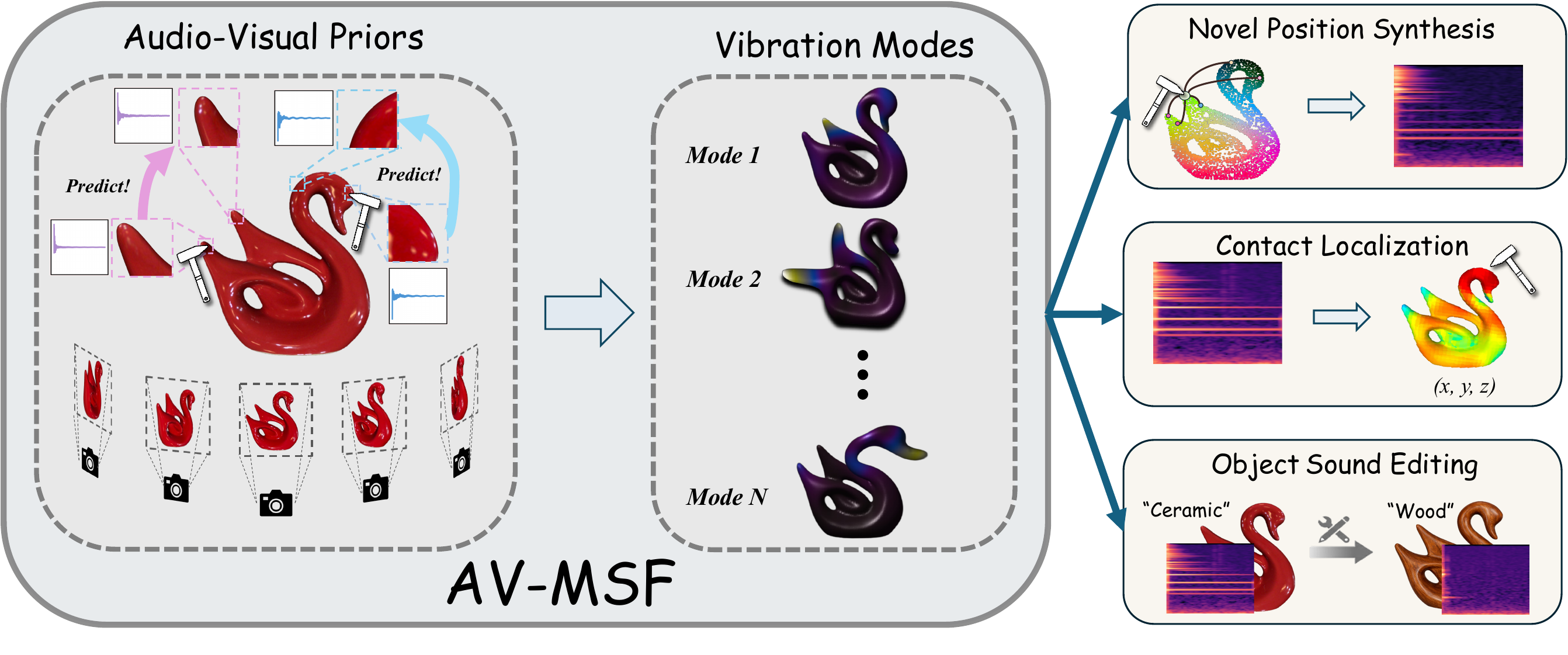}
  \caption{\emph{\textbf{Left: }}We reconstruct the Audio-Visual Modal Sound Field (\ourmethod) from multi-view RGB observations and only a few impact recordings, leveraging the insight that visually similar object regions exhibit similar vibration patterns. \emph{\textbf{Right: }}\ourmethod enables diverse applications, including novel-position impact sound synthesis, contact localization, and object sound editing.
  }
  \label{fig:teaser}
\end{figure}

Our everyday lives are filled with a wide variety of physical objects. From a quick glimpse, we can often infer an object's shape, appearance, and semantic function. Yet a full understanding of an object goes beyond passive visual observation: it often requires active physical interaction, which generates vibrations and sound. 
For example, a transparent wine goblet may look nearly identical whether it is made of glass or plastic. A light tap, however, makes the difference immediately clear: glass produces a bright, ringing sound, while plastic sounds duller and more heavily damped. In daily life, we rely on such impact sounds to judge material properties, stiffness, and thickness---physical attributes that are often hard to infer reliably from appearance alone.

Despite the importance of modeling objects' acoustics, most existing approaches to representing or virtualizing objects remain overwhelmingly vision-centric. 
Modern 3D reconstruction pipelines can recover detailed geometry and appearance from images, enabling downstream applications such as novel view synthesis~\cite{gausssplatting, mildenhall2020nerf}, controllable relighting~\cite{NeRFactor, RNG, IllumiNeRF, yu2023osf}, and text-driven 3D editing~\cite{ye2026nanod, sella2023vox, qi2024tailor3dcustomized3dassets}. Extending these representations to jointly model photorealistic appearance \emph{and} physically consistent impact sounds would unlock new capabilities, including more faithful and immersive contact feedback in VR, impact sound simulation for training multimodal embodied agents, and scalable creation of  realistic multisensory object assets.

Existing approaches to modeling object impact sounds broadly fall into two categories: \emph{physics-based} and \emph{data-driven}. In engineering and graphics, physics-based rigid-body impact sound synthesis is often performed via linear modal analysis~\cite{Timbrefields, Brien_SynthesizingSounds, FoleyAutomatic, ren2013example}. While physically interpretable, these simulations are often computationally expensive and can be inaccurate due to coarse or uncertain material parameters. Recent differentiable inverse-rendering pipelines seek to recover these physical parameters from real recordings via gradient-based optimization~\cite{diffsound,clarke2021diffimpact}; however, they either involve computationally heavy, error-prone optimization or adopt simplified modal parameterizations. In contrast, data-driven generative methods learn to synthesize impact sounds conditioned on multimodal cues~\cite{sonicgauss,su2023physics,SoundingthatObject}. Although they can produce diverse and perceptually plausible results, they are typically less physically grounded and often require substantial training data to generalize across objects and contact conditions.

To bridge this gap, our approach builds upon two key observations. First, impact sounds vary predictably across an object and are strongly correlated with local visual and geometric cues---how an object region looks often indicates how it will sound when struck. For example, as shown in \cref{fig:teaser}, tapping the two symmetric wings of a swan sculpture produces nearly identical vibrations, and the impact sound produced at the swan's neck can be inferred from the sound at neighboring locations with similar curvature. These visual priors can potentially help the model generalize even from only a few impact recordings. Second, an object's impact sound field can be distilled into a compact set of physically meaningful modal parameters. In particular, the model vibration frequencies are global properties determined by an object's shape and material, whereas each mode's  excitation gain is position-dependent, capturing how strongly that mode is activated at different contact positions. This representation is both physically grounded and more sample-efficient to learn.

Building on these insights, we introduce the \emph{Audio-Visual Modal Sound Field (\ourmethod)}, an object-level representation that reconstructs an object's modal impact-sound field from multi-view RGB observations and only a few impact recordings. \ourmethod first builds a 3D visual representation via 3D Gaussian Splatting (3DGS), and then lifts pre-trained visual features into a dense 3D feature field, providing a geometry-aware visual prior. Conditioned on this visual prior and sparse impact recordings, \ourmethod  reconstructs an object modal sound field parameterized by physically meaningful vibration modes, with additional residual components capturing unmodeled environmental effects. Together, \ourmethod offers an end-to-end framework that learns efficiently from real-world audio-visual data, while remaining physically interpretable.

Experiments on two real-world datasets, \textsc{ObjectFolder Real}~\cite{gao2023objectfolder} and \textsc{RealImpact}~\cite{clarke2023realimpact}, show that our approach significantly outperforms both physics-based and data-driven baselines. In the 20\% training-data setting, we obtain a $2\times$ improvement over both prior physics-based methods and data-driven baselines pretrained on large-scale simulated data. Moreover, our audio-visual, physics-based representation unlocks new downstream applications as illustrated in \cref{fig:teaser}. Because the representation includes position-dependent parameters that capture how an object responds differently across contact locations, \ourmethod  can infer the impact position from a previously unseen recording. Finally, by disentangling material-related acoustic parameters, our representation supports intuitive sound editing: we introduce an object sound editing pipeline that modifies an object's acoustic properties using Audio Score Distillation (ASD)~\cite{richter2025audiosds}.

Our key contributions are: (1) We present Audio-Visual Modal Sound Field (\ourmethod), a physics-based representation that reconstructs an object's modal sound field from multi-view RGB observations and a few impact sound recordings; (2) We show that \ourmethod substantially improves impact sound synthesis at novel contact locations over prior methods on two real-world datasets; (3) We demonstrate new downstream applications enabled by \ourmethod, including contact localization and object sound editing.
\section{Related Work}

\noindent\textbf{Virtualizing 3D Objects.}
Object digitization has long been a central problem in computer vision. Scanning-based methods recover high-fidelity meshes via laser scanning and range sensing~\cite{MichelangeloProject, rangescans}. In parallel, image-based reconstruction methods recover 3D representation from multi-view observations. While earlier systems~\cite{colmap, MVS} often produced coarse geometries, recent advances like NeRF~\cite{mildenhall2020nerf} and 3D Gaussian Splatting (3DGS)~\cite{gausssplatting} enable photorealistic reconstruction. These representations have powered a wide range of downstream tasks, including novel-view synthesis~\cite{mildenhall2020nerf,gausssplatting,Mip-NeRF}, relighting~\cite{NeRFactor,RNG,IllumiNeRF,yu2023osf}, geometry recovery~\cite{SuGaR, NeuS, Huang2DGS2024}, and controllable editing~\cite{ye2026nanod,sella2023vox,qi2024tailor3dcustomized3dassets}. Despite the progress, most existing pipelines remain largely appearance-centric; in contrast, we aim to incorporate position-dependent impact sound into the object digitization process.

\noindent\textbf{Multisensory Object Modeling.} 
Beyond visual appearance, a growing body of work begins to model additional object-centric sensory modalities, particularly audio and touch. Earlier efforts focused on capturing multisensory object properties to build interactive physical assets~\cite{scanningphysical}. More recently, audio and touch have been combined with vision to enable tasks such as 3D reconstruction~\cite{smith20203d,Suresh22icra}, cross-modal generation~\cite{zhang2017shape,tarf}, contact localization~\cite{lee2025sonicboom,luo2015localizing}, and robotic manipulation~\cite{li2022seehearfeel,higuera2025tactile}. Supporting these multisensory tasks, several new datasets have also been introduced. The ObjectFolder series~\cite{gao2021ObjectFolder,gao2022objectfolder} builds datasets of implicitly represented neural objects that encode visual, acoustic, and tactile observations, whereas VibraVerse~\cite{VibraVerse} explicitly models acoustics through modal analysis aligned with object geometry. However, both heavily rely on simulation and therefore suffer from sim-to-real gaps. Complementing these directions, ObjectFolder Real~\cite{gao2023objectfolder},  RealImpact~\cite{clarke2023realimpact}, and X-Capture\cite{clarke2025x} take important steps toward real-world multisensory capture, but their controlled acquisition setups limit scalability. Together, these limitations highlight the need for efficient reconstruction pipelines that can scale the creation of realistic multisensory assets.

\noindent\textbf{Physics-based Impact Sound Synthesis.}
Traditional physics-based methods~\cite{Timbrefields, Brien_SynthesizingSounds, FoleyAutomatic, ren2013example} synthesize impact sounds by modeling the vibration modes of rigid objects. This is typically achieved through modal analysis, where sound is generated by solving the wave equation given a volumetric mesh and material properties such as Young's modulus, Poisson's ratio, and density. More recent differentiable frameworks~\cite{diffsound, clarke2021diffimpact} extend this paradigm toward inverse rendering of physically meaningful acoustic parameters from recorded audio. DiffSound~\cite{diffsound} estimates physical and shape attributes through a differentiable pipeline. While physically interpretable, it is computationally expensive and susceptible to error accumulation due to its long optimization chain involving high-order finite element methods (FEM). In contrast, DiffImpact~\cite{clarke2021diffimpact} improves efficiency by parameterizing impact sounds with modal quantities such as frequency, damping, and gain instead of explicitly optimizing material properties. Building on these works, we aim to reconstruct a physically interpretable modal sound field in a more efficient and practical regime. By incorporating visual priors, our method avoids costly FEM-based forward simulation over volumetric meshes while preserving the few-shot and physically grounded benefits of inverse approaches.

\noindent\textbf{Data-driven Impact Sound Synthesis.}
Data-driven methods synthesize impact sounds by learning directly from real-world recordings with deep neural networks (often conditioned on multimodal cues). Video-to-sound approaches~\cite{Owens_2016_CVPR, su2023physics, dou2025hearing} can generate semantically plausible Foley effects, but do not model an explicit 3D acoustic representation. More recent works~\cite{li2025visualacousticfields,sonicgauss} move toward 3D audio-visual modeling by embedding acoustics into 3D object or scene representations. However, these methods rely on fine-tuning pre-trained text-to-audio models, whose priors are often mismatched to transient, contact-driven impact sounds. Overall, purely data-driven pipelines tend to be task-specific and lack physical interpretability; as a result, they typically require large-scale training data---often difficult to obtain---to generalize reliably, which limits their applicability to diverse downstream applications.
\section{Approach}

\begin{figure}[t]
\centering
\includegraphics[width=1\linewidth]{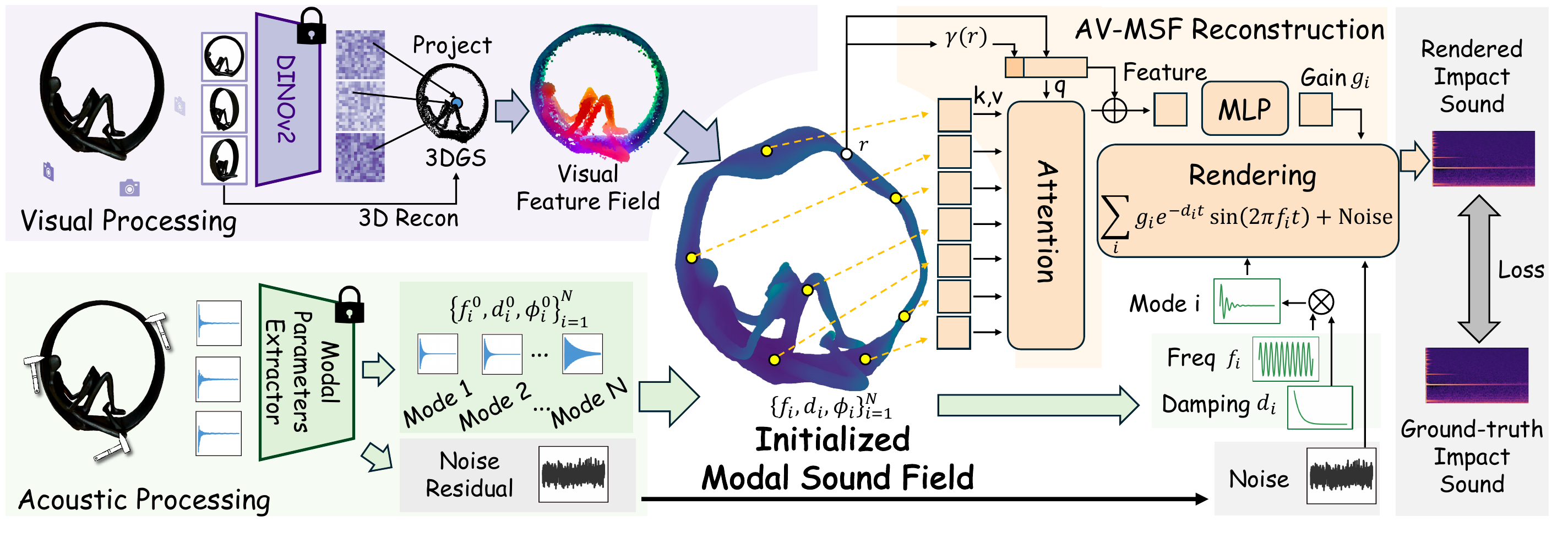}
\caption{\textbf{Overview of \ourmethod.} Given multi-view image observations and few-shot impact recordings, our framework reconstructs an audio-visual modal sound field. \textbf{(1) Visual Processing:} We extract dense features with a pre-trained vision encoder and lift them into a 3D Gaussian Splatting (3DGS) representation to form a geometry-aware visual feature field. \textbf{(2) Acoustic Processing:} We extract modal parameters from a few reference recordings to initialize optimization, and model unmodeled environmental noise with a residual component. \textbf{(3) \ourmethod Reconstruction:} We jointly optimize global modal frequencies and dampings, residual noise, and an implicit spatially varying neural gain field, guided by the visual feature field. 
}
\label{fig:method}
\end{figure}

\vspace{-0.01in}

We first review the background on modal sound synthesis in \cref{sec:background} and define our task and relevant notations in \cref{sec:problem_formulation}. Next, we describe the visual processing and acoustic processing components in \cref{sec:visual_processing} and \cref{sec:sound_modeling}, respectively. We then discuss our \ourmethod reconstruction pipeline in \cref{sec:MSF_reconstruction}. Finally, we introduce two downstream applications enabled by our method in \cref{sec:applications}. Our method is illustrated in \cref{fig:method}.

\subsection{Background on Modal Sound Synthesis}
\label{sec:background}
We first briefly review \emph{linear modal analysis}~\cite{JamesSIGGRAPHCourses}, the standard pipeline to synthesize rigid-body sounds from an object's physical properties and shape.

\noindent\textbf{Elastic Vibration from Physical Parameters.}
Given an object with volumetric mesh and material properties, linear modal analysis models its small elastic vibration as:
\begin{equation}
    \mathbf{M}\ddot{\mathbf{u}}(t) + \mathbf{D}\dot{\mathbf{u}}(t) + \mathbf{K}\mathbf{u}(t) = \mathbf{f}(t),
    \label{eq:elastodynamics}
\end{equation}
where $\mathbf{u}(t)$ denotes the nodal displacement, $\mathbf{M}$ and $\mathbf{K}$ are the mass and stiffness matrices, $\mathbf{D}=\alpha \mathbf{M} + \beta \mathbf{K}$ is the Rayleigh damping, and $\mathbf{f}(t)$ is the external force. Here, $\mathbf{M}$ depends on the object's shape and density $\rho$, while $\mathbf{K}$ depends on its shape and material parameters (Young's modulus $E$ and Poisson's ratio $\nu$).

\noindent\textbf{Modal Decomposition and Sound Synthesis.}
By generalized eigenanalysis $\mathbf{K}\mathbf{U}=\mathbf{M}\mathbf{U}\mathbf{S}$, Eq.~\eqref{eq:elastodynamics} can be reformulated as:
\begin{equation}
    \ddot{\mathbf{q}}(t) + (\alpha \mathbf{I} + \beta \mathbf{S})\dot{\mathbf{q}}(t) + \mathbf{S}\mathbf{q}(t) = \mathbf{U}^\top \mathbf{f}(t),
\end{equation}
where $\mathbf{U}$ is the matrix of vibration modes, $\mathbf{S}$ is the diagonal matrix of modal eigenvalues, and $\mathbf{q}(t)$ satisfies $\mathbf{u}(t)=\mathbf{U}\mathbf{q}(t)$, with each entry describing the response of one mode over time. The resulting impact sound is modeled as the sum of all decaying modal vibrations:
\begin{equation}
    s(t)=\sum_{i=1}^{N} g_i e^{-d_i t}\sin(2\pi f_i t),
    \label{eq:sinusoids}
\end{equation}
where $f_i$, $d_i$, $g_i$ are the frequency, damping, gain of the $i$-th mode, respectively.

\paragraph{Assumptions and Limitations.}
Notably, our approach relies on two main assumptions. First, we assume the modal frequency $f_i$ and damping $d_i$ are intrinsic, position-invariant properties of the object, while only the modal gain $g_i$ varies with contact location. Second, the Rayleigh damping model of linear modal analysis is a numerical approximation of real physical behaviors~\cite{JamesSIGGRAPHCourses}, which do not hold for all real-world objects. As we observe in real datasets, the damping can vary across contact locations for some objects. We therefore make damping optionally learnable as a spatially varying parameter. For simplicity, however, we denote only the gain as a spatially varying parameter in our formulation below.

\subsection{Problem Formulation}
\label{sec:problem_formulation}
We consider the problem of reconstructing an object's Audio-Visual Modal Sound Field (\ourmethod) from visual observations and few-shot impact sound recordings.  Specifically, for a single object, we are given a set of calibrated multi-view RGB images $\mathcal{I}=\{(I_i, \Pi_i)\}_{i=1}^I$, where $I_i$ denotes the $i$-th image and $\Pi_i$ denotes the known camera parameters, together with few-shot impact sound recordings $\mathcal{S}=\{(\mathbf{x}_j, s_j(t))\}_{j=1}^S,$ where $\mathbf{x}_j \in \mathbb{R}^3$ is the 3D contact location on the object surface and $s_j(t)$ is the corresponding audio waveform.

Our goal is to learn an object-level representation that can render the impact sound at an arbitrary novel contact location $\mathbf{x}$ on the surface. Following \cref{eq:sinusoids}, we model the impact sound generated at $\mathbf{x}$ as a combination of decaying modal vibrations (assuming the object is initially static) and a non-modal residual component:
\begin{equation}
s(\mathbf{x}, t) = \sum_{i=1}^{N} g_i(\mathbf{x}) \, e^{-d_i t}\sin(2\pi f_i t) + \sum_{i=1}^{F} \epsilon(m_i,t),
\label{eq:problem_modal_field}
\end{equation}
where $N$ is the number of modes, $\{f_i\}_{i=1}^{N}$ and $\{d_i\}_{i=1}^{N}$ are the modal frequencies and dampings, $g_i(\mathbf{x})$ is the excitation gain of the $i$-th mode at contact location $\mathbf{x}$, and $\epsilon(m_i,t)$ represents residual noise parameterized by a learnable frequency-domain magnitude $\{m_i\}_{i=1}^{F}$ over $F$ frequency bins.

As noted in \cref{sec:background}, $\{f_i\}_{i=1}^{N}$ and $\{d_i\}_{i=1}^{N}$ are object-intrinsic global parameters, while only the modal gains vary spatially across contact locations. We therefore represent the object's modal sound field as $\mathcal{F}=\Big(\{f_i, d_i\}_{i=1}^{N},\; \{m_i\}_{i=1}^{F},\; \mathcal{G}_\theta(\mathbf{x})\Big),$ where $\mathcal{G}_\theta(\mathbf{x}) = [g_1(\mathbf{x}), \dots, g_N(\mathbf{x})]^\top$ is an implicit spatial gain field that predicts modal gains for any queried surface point $\mathbf{x}$.

Overall, the reconstruction problem can thus be formulated as a mapping $\Phi : (\mathcal{I}, \mathcal{S}) \mapsto \mathcal{F},$ which estimates (i) a set of global modal parameters $\{f_i,d_i\}_{i=1}^{N}$ and noise residual magnitudes $\{m_i\}_{i=1}^{F}$, and (ii) an implicit neural gain field $\mathcal{G}_\theta(\mathbf{x})$ that predicts the modal gains for any queried surface location $\mathbf{x}$.

\subsection{Multi-View Visual Feature Extraction and Refinement}
\label{sec:visual_processing}
To effectively guide acoustic prediction of impact sound, we construct a geometry-aware 3D feature field from multi-view RGB images. We first reconstruct a 3D Gaussian Splatting (3DGS) representation, yielding a dense point cloud of Gaussian centers $\mathcal{P} = \{\mathbf{o}_i\}_{i=1}^O$. We then extract 2D visual embeddings using pre-trained DINOv2~\cite{oquab2024dinov} and lift them into 3D, assigning each Gaussian spatial center $\mathbf{o}_i$ a feature vector $\mathbf{f}_i \in \mathbb{R}^D$.

However, due to viewpoint variations, these lifted features may fail to respect the object's intrinsic geometric symmetries. For example, points lying on the same rotational orbit should ideally share similar semantics. To address this issue and enforce geometric consistency, we automatically detect object symmetries, such as rotational or planar mirror symmetries, by evaluating the geometric alignment error of candidate transformations over the point cloud. Given the detected symmetries, we perform symmetry-aware feature alignment. Specifically, we aggregate and average the semantic features across their corresponding symmetric counterparts through orbit or reflection pooling, explicitly enforcing geometric consistency and producing  final refined feature $\mathbf{f}_i^{\text{refined}}$ for each Gaussian. For simplicity, we use $\mathbf{f}_i$ in the following. See Supp. for details on the feature refinement procedures.

\subsection{Modal Sound Parameters Extraction and Initialization}
\label{sec:sound_modeling}
Directly optimizing $\mathcal{F}$ from few-shot recordings $\mathcal{S}$ is highly non-convex and prone to local minima. To stabilize training, we leverage the physical prior that  the modal frequencies $\{f_i\}_{i=1}^{N}$ and dampings $\{d_i\}_{i=1}^{N}$ are intrinsic object properties, explicitly extract them from $\mathcal{S}$ for initialization.

\vspace{0.01in}

\noindent\textbf{Modal Parameters Extraction.} For each recorded impact sound $s_j(t) \in \mathcal{S}$, we compute its Short-Time Fourier Transform (STFT) and apply a per-frequency inverse-STFT to decompose the audio into a set of narrow-band time-domain signals, $\{s_{j,n}(t)\}_{n=1}^{F}$, where $F$ is the total number of frequency bins. We explicitly filter out noise by discarding peaks that occur beyond a time threshold or are not significantly damped. For each candidate mode signal $s_{j,n}(t)$, we estimate its damping $d_n$ and gain $g_n$ via log-linear regression:
\begin{equation}
\log |\mathcal{H}\{s_{j,n}(t)\}| = \log(g_n) - d_n \cdot t,
\end{equation}
where $\mathcal{H}$ denotes the Hilbert transform. To obtain the global object-level parameters, we identify the frequencies that consistently appear across recordings. For the $i$-th selected mode, the global damping coefficient $d_i$ is computed as the average of the estimated dampings across all recordings. These globally extracted $\{f_i, d_i\}_{i=1}^{N}$ serve as the shared intrinsic parameters for the modal sound field.

\vspace{0.01in}

\noindent\textbf{Residual Noise Modeling.} Real-world sound recordings often contain non-modal effects. To stabilize modal parameter optimization, we model a residual component to capture (1) low-frequency background noise and (2) other unpredictable non-modal factors, such as variations in contact force and microphone distance. We model the residual component as static filtered noise with learnable per-band magnitudes $\mathbf{m}=\{m_i\}_{i=1}^{F}$ following prior work~\cite{clarke2021diffimpact}. We initialize $\{m_i\}_{i=1}^{F}$ from $\mathcal{S}$ by averaging the lowest-energy temporal segments within each frequency bin. At synthesis time, we draw white noise $\epsilon(t)\sim\mathcal{N}(0,1)$ and shape it with a differentiable frequency filter:
\begin{equation}
\epsilon(\mathbf{m},t)=\sum_{i=1}^{F} m_i \,\big(b_i * \epsilon(t)\big),
\end{equation}
where $b_i$ denotes the band-pass filter for the $i$-th frequency bin. The final waveform is obtained by adding this filtered residual to the modal reconstruction. 

\subsection{Reconstructing Audio-Visual Modal Sound Field}
\label{sec:MSF_reconstruction}
Up to this point, we have obtained a geometry-aware 3DGS representation and initialized with extracted acoustic parameters $\{f_i, d_i\}_{i=1}^{N}$ and $\{m_i\}_{i=1}^{F}$ (middle of~\cref{fig:method}). Our goal is to leverage the visual prior to optimize $\mathcal{F}$ through the differentiable modal sound synthesizer in \cref{eq:problem_modal_field}.

For each object, we cluster the Gaussian centers $\mathcal{P}=\{\mathbf{o}_i\}_{i=1}^{O}$ into $K$ spatial groups using 3D Euclidean distance, with $\mathbf{c}_k$ denoting the center of cluster $k$. Within each cluster, we average the symmetry-aligned features to form a single region-level descriptor:
\begin{equation}
\bar{\mathbf{f}}_k \;=\; \frac{1}{|\mathcal{C}_k|}\sum_{i\in\mathcal{C}_k}\mathbf{f}_i,
\end{equation}
where $\mathcal{C}_k$ denotes the set of Gaussians assigned to cluster $k$. The resulting descriptors $\{\bar{\mathbf{f}}_k\}_{k=1}^{K}$ provide a compact set of visual priors.

\vspace{0.01in}

\noindent\textbf{Predicting Location-dependent Gains.}
Given a queried impact location $\mathbf{x}$, we first find its nearest Gaussian center $i^\star = \arg\min_{i}\|\mathbf{x}-\mathbf{x}_i\|_2$, and use its local visual feature $\mathbf{f}_{i^\star}$ as a fine-grained cue. We also encode geometric context by computing the relative offsets from the query point to all cluster centers. Specifically, $r_k = \mathbf{x}_{i^\star}-\mathbf{c}_{k}.$ and we stack the offsets to all K clusters into $\mathbf{r} = [r_1,\dots,r_K]$.

We then predict the modal gains $\mathcal{G}_\theta(\mathbf{x}) = [g_1(\mathbf{x}),\dots,g_N(\mathbf{x})]^\top$ by conditioning on (i) the local feature $\mathbf{f}_{i^\star}$, (ii) the offset encoding $\mathrm{PE}(\mathbf{r})$, and (iii) global context aggregated from all $K$ cluster descriptors via attention. Specifically, we follow NeRF~\cite{mildenhall2020nerf} and use sin-cos function as our $\mathrm{PE}(\cdot)$ function.
Concretely, we form a query vector
\begin{equation}
\mathbf{q} = W_q\,[\mathrm{PE}(\mathbf{r});\mathbf{f}_{i^\star}],
\end{equation}
and define keys/values from the region descriptors
\begin{equation}
\mathbf{k}_k=W_k\bar{\mathbf{f}}_k,\qquad \mathbf{v}_k=W_v\bar{\mathbf{f}}_k.
\end{equation}
Dot-product attention produces a context feature
\begin{equation}
\alpha_k = \mathrm{softmax}_k\!\Big(\frac{\mathbf{k}_k^\top \mathbf{q}}{\sqrt{d}}\Big),\qquad
\mathbf{z}=\sum_{k=1}^{K}\alpha_k\,\mathbf{v}_k.
\end{equation}
Finally, the context feature is projected and concatenated with the local Gaussian feature to predict the modal gains:
\begin{equation}
\mathcal{G}_\theta(\mathbf{x}) = \mathrm{MLP}_\theta(\mathbf{z}; \mathbf{f}_{i^\star}).
\end{equation}

\noindent\textbf{Differentiable Synthesis and Optimization.}
For each observed impact recording $(\mathbf{x}_j,s_j(t))\in\mathcal{S}$, we synthesize
\begin{equation}
\hat{s}_j(t)=\sum_{i=1}^{N} \mathcal{G}_\theta(\mathbf{x}_j)\,e^{-d_i t}\sin(2\pi f_i t) \;+\; \epsilon(\mathbf{m},t),
\end{equation}
where $\epsilon(\mathbf{m},t)$ is the filtered-noise residual from \cref{sec:sound_modeling}.
We optimize $\theta$ jointly with the global parameters $\{f_i,d_i\}_{i=1}^{N}$ and $\mathbf{m}$ by minimizing a reconstruction loss between $\hat{s}_j(t)$ and $s_j(t)$ over few-shot contacts in $\mathcal{S}$.

\vspace{0.01in}

\noindent\textbf{{Training Objectives.}} We train each object using a two-stage optimization strategy. In the warm-up stage, we optimize $\mathcal{G}_\theta$ using the gain field extracted in \cref{sec:sound_modeling}. For each audio sample $i$, we predict the gain $\mathcal{G}_\theta(x_i)$ and minimize the mean-squared error (MSE) with respect to the extracted target gain $\mathbf{g}_i$:
\begin{equation}
\mathcal{L}_{\text{warmup}} =
\frac{1}{S}\sum_{i=1}^{S}
\left\|
\mathcal{G}_\theta(\mathbf{x}_i)-\mathbf{g}_i
\right\|_2^2 .
\end{equation}
This warm-up anchors the gain field in a well-conditioned optimization regime, reducing the risk of collapse and accelerating convergence.

In the training stage, we train all parameters end-to-end with a multi-scale STFT reconstruction loss. We apply it to both the full waveform and the early-time segment of length $t$, where the impact energy is most concentrated and the signal is less affected by environmental noise. The training loss is defined as:
\begin{equation}
\mathcal{L}_{\text{spec}} =
\sum_{r}
\left(
\left\|
\mathrm{STFT}_r(s_i) - \mathrm{STFT}_r(\hat{s}_i)
\right\|_1
+
w_c
\left\|
\mathrm{STFT}_r\!\left(s_i[:t]\right) - \mathrm{STFT}_r\!\left(\hat{s}_i[:t]\right)
\right\|_1
\right),
\end{equation}
where $r$ indexes FFT resolutions and $w_c$ is the weight for the early cut-off loss.

\subsection{Applications}
\label{sec:applications}
Our physics-based audio-visual modal sound field can potentially support a range of downstream tasks by leveraging the distinct physical roles of its components. We showcase two below: contact localization and object sound editing. 

\noindent\textbf{Contact Localization.}
In this task, we aim to infer the 3D impact location $\mathbf{x}^*$ from a novel sound recording $s_{new}(t)$. Relying on the object-intrinsic global parameters $\{f_i, d_i\}_{i=1}^{N}$ obtained during training, we first extract the corresponding mode-specific gains $\hat{\mathbf{g}} = [\hat{g}_1, \dots, \hat{g}_N]^\top$ from $s_{new}(t)$. Because our learned spatial gain field $\mathcal{G}_\theta(\mathbf{x})$ uniquely characterizes the acoustic response at any point on the object, we can localize the impact by finding the surface coordinate $\mathbf{x}$ that minimizes the discrepancy between the predicted and extracted gains:
\begin{equation}
\mathbf{x}^* = \arg\min_{\mathbf{x}} \mathcal{D} \left( \mathcal{G}_\theta(\mathbf{x}), \hat{\mathbf{g}} \right),
\end{equation}
where $\mathcal{D}$ denotes a distance metric. In practice, we use cosine distance to ensure the localization is invariant to the absolute magnitude of the initial impact force.

\noindent\textbf{Object Sound Editing.}
To enable semantic, text-driven editing of the impact sound, we adopt Audio-SDS~\cite{richter2025audiosds}. Given a target text prompt $p$, we optimize the object-intrinsic modal parameters $\theta = \{f_i, d_i, g_i\}_{i=1}^N$ by rendering the modal audio $s(\theta, t)$ and distilling guidance from a frozen, pretrained text-to-audio diffusion model. A significant challenge arises from the fact that modal frequencies are sparse, in which independent optimization of $\{f_i\}_{i=1}^N$ often struggles to overcome the large spectral gap between contrastive materials (e.g., shifting from wood to metal). To facilitate convergence, we propose a hierarchical frequency parameterization: $f_i = \exp(\log \alpha + \log f_{i, \text{init}} + \Delta f_i)$. Here, $\alpha$ serves as a global pitch-scaling factor that captures the primary material-dependent shift, while $\Delta f_i$ allows for per-mode residual refinement. 

To circumvent optimization instabilities, we employ Decoder-SDS combined with multi-step denoising. Specifically, we map the rendered audio to the latent space, add noise, perform partial DDIM sampling steps conditioned on $p$, and decode the result back to the audio domain to form a target pseudo-ground-truth waveform $\hat{s}(t)$. The parameters $\theta$ are updated by minimizing a multi-resolution STFT loss between the rendered $s(\theta, t)$ and the distilled target $\hat{s}(t)$.

\paragraph{Physically Grounded Sound Editing.}
While Audio-SDS successfully modifies the acoustic parameters $\{f_i, d_i\}_{i=1}^{N}$ and reference gains $\{g_i\}_{i=1}^{N}$ to match a text prompt, we must ensure these edits remain consistent with linear modal analysis to preserve the object's acoustic physics. As detailed in Supp., altering Young's modulus, density, and Rayleigh damping coefficients scales the modal frequencies and decay rates but leaves the spatial mode shapes unchanged. Consequently, after editing material parameters, the modal gains update via a simple per-mode rescaling $a_k^{(2)}(x) = a_k^{(1)}(x)\frac{f_k^{(1)}}{f_k^{(2)}}.$
Because the frequency-dependent factors cancel out, the gain ratio between any two impact locations remains invariant to such parameter editing. This physical property allows us to seamlessly preserve the learned spatial modal field and perform physically grounded sound editing by only updating the per-mode frequencies, decay rates, and gain scales.
\section{Experiment}
We now validate \ourmethod on novel position object impact sound rendering and two new downstream applications, and compare it against prior approaches and baseline methods.

\subsection{Datasets}
We evaluate our method on \textsc{ObjectFolder Real}~\cite{gao2023objectfolder} and \textsc{RealImpact}~\cite{clarke2023realimpact}, two real-world multisensory datasets that include impact sound recordings. For both datasets, we normalize the impact sound recordings using the provided force profiles, zero-align the peak signals, and truncate them to 3 seconds to preserve the full damping behavior. Dataset-specific details are provided below.

\noindent\textbf{\textsc{ObjectFolder Real~\cite{gao2023objectfolder}}} contains multisensory measurements of 100 real-world objects across seven material types, including 3D meshes, videos, impact sounds, and tactile readings. Each object is associated with 30--50 impact sounds manually recorded across its surface using a muted, force-sensing impact hammer. By default, 20\% of the impact recordings for each object are selected via farthest-point sampling for training, with the rest used for evaluation. We use the dataset for the main quantitative comparison of novel-position impact sound synthesis, and its subset for ablations and downstream evaluations.

\noindent\textbf{\textsc{RealImpact~\cite{clarke2023realimpact}}} contains 150,000 impact sound recordings from 50 everyday objects. Because it was designed to study spatial acoustics of impact sounds, each object has only 5 distinct impact locations, each struck by an automatically controlled, calibrated impact hammer mounted on a rotational gantry system, and recorded 600 times from different microphone positions. In our experiments, we use only the  microphone closest to the object, resulting in 5 recordings per object. Due to its limited number of impact locations, we adopt a leave-one-out cross-validation protocol, iteratively evaluating on one impact location while training on the remaining ones.

\subsection{Baselines and Evaluation Metrics}
We compare to existing physics-based and data driven methods~\cite{diffsound,sonicgauss} and multiple baselines on novel-position impact sound rendering:

\noindent\textbf{\textsc{White Noise}:} Random noise scaled to match the average loudness of the impact recordings used to train our method.

\noindent\textbf{\textsc{Random Impact}:} We use a randomly selected impact sound from another object as the predicted impact sound for the queried location and object.

\noindent\textbf{\textsc{KNN:}} For each query contact location, we retrieve the K nearest training impact sounds and average them to produce the rendered impact sound for the queried location. We use K=3 in all of our settings.

\noindent\textbf{\textsc{DiffSound}~\cite{diffsound}} is a state-of-the-art physics-based method that estimates material parameters via inverse rendering. Position-specific impact sounds can be simulated from the object's vibration modes using the inferred parameters. We further add a learned noise residual from our method for fairness. 

\noindent\textbf{\textsc{SonicGauss}~\cite{sonicgauss}} is a state-of-the-art data-driven method that trains an impact sound diffusion model conditioned on  contact position and encoded 3D Gaussian object features. We finetune each object from its Stage-2 checkpoint on \textsc{ObjectFolder 2.0}, with improved force and audio normalization.

\paragraph{Evaluation Metrics.} We compare with baselines using following metrics: (1) \emph{L1 Distance}, which computes the Euclidean distance between the ground-truth and predicted spectrograms; (2) \emph{L1-log Distance}, which is the same as L1 Distance but measuring the distance between log-mel spectrograms for better perceptual alignment; (3) \emph{Envelope Distance}, which computes the Euclidean distance between the amplitude envelopes of the ground-truth and the predicted waveforms; and (4) \emph{CDPAM}~\cite{CDPAM}, a perceptual audio similarity metric; (5) \emph{Relative Mean Euclidean Distance (RMED)}, distance between the predicted and ground-truth coordinates normalized by the bounding-box diagonal for contact localization.

\begin{table*}[t!]
\centering
\small
\caption{\textbf{Quantitative comparison on novel-position impact sound rendering.}  We compare all the baselines on two popular real-world impact sound datasets, \textsc{ObjectFolder Real}~\cite{gao2023objectfolder} and \textsc{RealImpact}~\cite{clarke2023realimpact}. Lower is better for all metrics.
}
\label{tab:nps}
\begin{tabular}{lcccccccc}
\toprule
& \multicolumn{4}{c|}{\textsc{ObjectFolder Real}} & \multicolumn{4}{c}{\textsc{RealImpact}} \\
\cmidrule(lr){2-5}\cmidrule(lr){6-9}
Method 
 & L1 & L1 Log & ENV  & CDPAM
 & L1 & L1 Log & ENV  & CDPAM \\
\midrule

\textsc{White Noise}
& 3.774 & 6.859 & 0.305 & 1.35e-3 & 3.789 & 7.258 & 0.308 & 1.23e-3 \\
\textsc{Random Impact}
& 0.035 & 1.367 & 0.019 & 2.47e-4 & 0.039 & 1.582 & 0.022 & 2.71e-4 \\
\textsc{KNN}
& 0.014 & \textbf{0.930} & 0.014 & 1.53e-4 & 0.026 & 1.036 & 0.017 & 2.25e-4 \\
\textsc{DiffSound}~\cite{diffsound}
& 0.031 & 1.298 & 0.030 & 2.53e-4 & 0.029 & 1.533 & 0.024 & 2.39e-4 \\
\textsc{SonicGauss}~\cite{sonicgauss}
& 0.033 & 1.281 & 0.018 & 2.01e-4 & 0.039 & 1.580 & 0.025 & 2.43e-4 \\
\textbf{Ours}
& \textbf{0.013} & 0.951 & \textbf{0.014} & \textbf{1.35e-4} & \textbf{0.021} & \textbf{0.996} & \textbf{0.017} & \textbf{2.16e-4} \\

\bottomrule
\end{tabular}
\end{table*}

\subsection{Novel Position Impact Sound Rendering}
We report novel position impact sound rendering results in Tab.~\ref{tab:nps}. \ourmethod substantially outperforms prior physics-based and data-driven baselines (\textsc{DiffSound} and \textsc{SonicGauss}). \textsc{DiffSound} underperforms because it heavily relies on inverse rendering to estimate material parameters from real recordings, which is often error-prone and leads to inaccurate rendered sound. \textsc{SonicGauss} is highly data-hungry: even with pretraining on \textsc{ObjectFolder 2.0}~\cite{gao2022objectfolder}, few-shot fine-tuning remains ineffective. Moreover, as a generative model, it tends to introduce hallucinated or distorted patterns, indicating difficulty in faithfully preserving the underlying physical properties. \textsc{KNN} is a strong baseline because many objects are symmetric (see object visualizations in Supp.), and nearby points often share similar spectra. Nonetheless, \ourmethod achieves better results on most metrics, particularly when local geometry or appearance changes sharply.

\paragraph{Case Study on Non-symmetric Objects.}
The evaluation datasets are mainly composed by symmetric objects (\eg, bowls, plates), which make KNN a strong baseline. To further demonstrate the strength of our method, we evaluate on a subset of less symmetric objects from ObjectFolder Real (see Supp. for more details). As shown in Tab.~\ref{tab:nonsym}, our method consistently outperforms the KNN baseline on this non-symmetric subset, with clear improvements across all metrics.

\begin{figure}[t!]
\centering
\includegraphics[width=\linewidth]{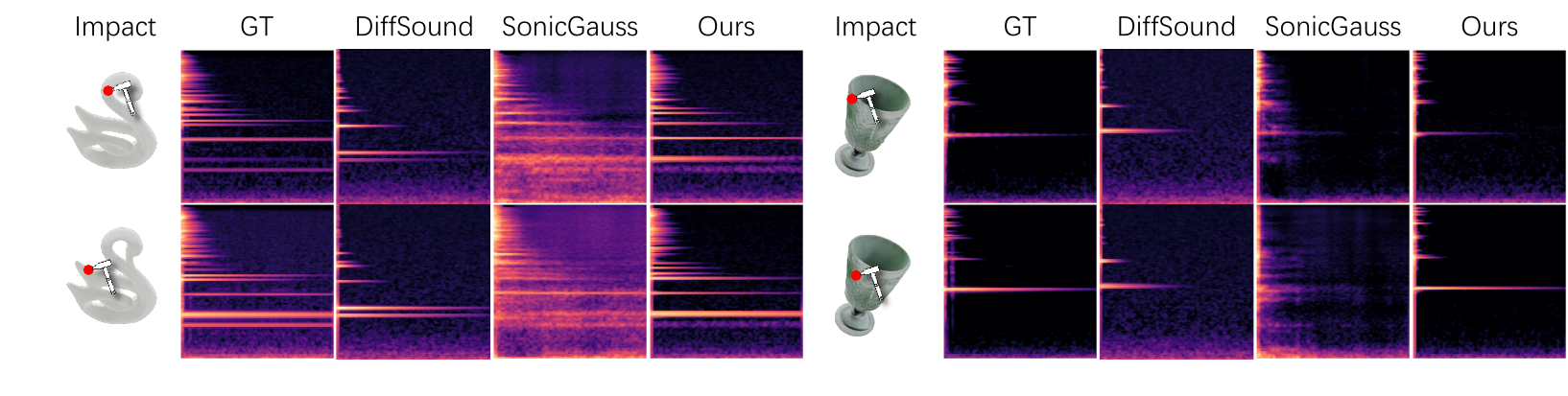}
\caption{\textbf{Qualitative Spectrogram Comparisons.} We visualize the rendered impact sounds from \textsc{DiffSound}~\cite{diffsound}, \textsc{SonicGauss}~\cite{sonicgauss}, and our method, alongside the ground truth. All spectrograms are generated on the settings of 20–20k Hz, the first second, scaled to $(-1,1)$. See Supp. for more examples.} 
\label{fig:compare}
\end{figure}

\paragraph{Qualitative Comparison.} Figure~\ref{fig:compare} presents qualitative comparisons on two objects, visualizing the 20–20k Hz spectrograms of the first second for two different impact locations per object. \textsc{DiffSound} produces clear and sharp structures, but the error in its physical parameter estimation can lead to incorrectly placed frequency bands or dampings. \textsc{SonicGauss} recovers the overall frequency content more roughly correctly, yet it can introduce hallucinated or distorted patterns due to its generative nature. In contrast, our method yields clearer, more faithful spectral structures while remaining strongly position-aware, producing distinct responses across impact locations that better match the ground truth.

\begin{table*}[t!]
\centering

\begin{minipage}[t]{0.48\textwidth}
\centering
\caption{Ablation Study Results.}
\label{tab:ablate}
\resizebox{\linewidth}{!}{
\begin{tabular}{lcccc}
\toprule
Method & L1 & L1 Log & ENV & CDPAM \\
\midrule
w/o visual & 0.028 & 1.148 & 0.015 & 4.20e-4 \\
w/o init & 0.045 & 1.077 & 0.026 & 2.97e-4 \\
w/o align & 0.019 & 1.002 & 0.016 & 2.62e-4 \\
w/o residual & 0.031 & 5.764 & 0.026 & 3.22e-4 \\
\midrule
ours & \textbf{0.019} & \textbf{0.927} & \textbf{0.015} & \textbf{2.00e-4} \\
\bottomrule
\end{tabular}
}
\end{minipage}
\hfill
\begin{minipage}[t]{0.48\textwidth}
\centering
\caption{Case Study Results on Non-symmetric Objects.}
\label{tab:nonsym}
\resizebox{\linewidth}{!}{
\begin{tabular}{lcccc}
\toprule
Method & L1 & L1 Log & ENV & CDPAM \\
\midrule
KNN & 0.0135 & 0.9724 & 0.0138 & 2.15e-4 \\
Ours & 0.0110 & 0.9259 & 0.0131 & 1.53e-4 \\
\bottomrule
\end{tabular}
}
\end{minipage}

\end{table*}

\paragraph{Ablation Study.}
We conduct an ablation study to assess the contributions of four key components in our method. The results, summarized in Tab.~\ref{tab:ablate}, confirm the effectiveness of each component. (1) \emph{Ablate on Appearance}, removing the DINO features, which provide local appearance, consistently degrades performance, with the largest drop on perceptual quality. We attribute this to DINO encoding fine-grained and semantically meaningful local details, such as curvature, edges, and texture. In contrast, 3D point coordinates mainly capture coarse geometry and therefore provide a weaker signal for accurate sound rendering. (2) \emph{Ablate on Modal Parameter Initialization Warm-up}, we further verify the importance of the modal parameter initialization, which yields a significant performance gain. This warm-up stage stabilizes optimization by bringing modal parameters into a physically reasonable region. Without it, training can collapse, as the model may become stuck in a poor local minimum. (3) \emph{Ablate on Feature Alignment}, directly lifting DINO features into 3D can introduce misalignment with the geometry prior due to lighting and texture variations; for instance, symmetric regions may receive inconsistent features across two sides. Our feature-alignment module mitigates this issue by enforcing geometric consistency, improving performance across all metrics. (4) \emph{Ablation on Residual Component}, removing the residual component yields the worst L1 Log performance, highlighting its importance for capturing low-frequency environmental noise. Moreover, removing it worsens the contact localization error from 38.4\% to 43.8\% on the ablation subset, suggesting that the gain field is otherwise sacrificed to fit non-modal effects.

\subsection{Results on Downstream Applications}

\begin{figure*}[t!]
\centering
\begin{minipage}[t]{0.48\textwidth}
\centering
\includegraphics[width=\linewidth]{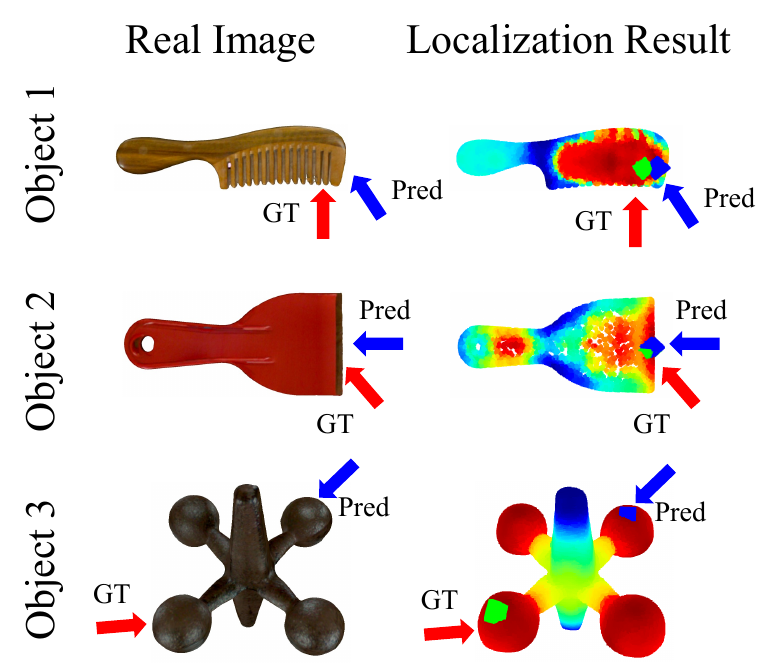}
\end{minipage}
\hfill
\begin{minipage}[t]{0.48\textwidth}
\centering
\includegraphics[width=\linewidth]{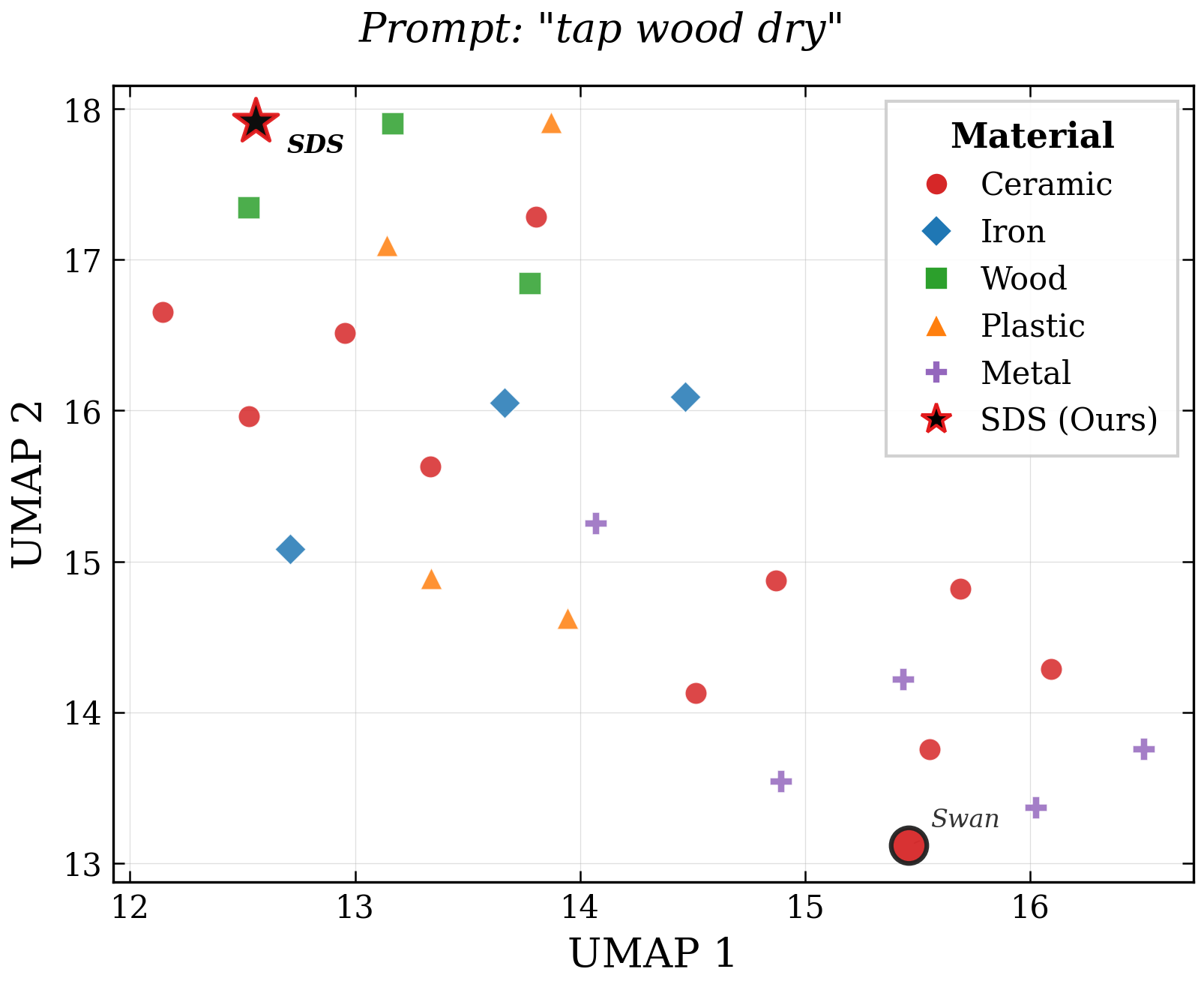}
\end{minipage}
\caption{\textbf{Examples for Downstream Applications.}  \textbf{Left}: Contact localization heatmaps. Red arrows indicate novel impact locations, and blue arrows indicate predictions. Object 3 shows a failure case. \textbf{Right}: Visualization of frequency distributions of modal parameters before and after sound editing.}
\label{fig:contact_sound_editing_visualization}
\end{figure*}

\begin{table*}[t!]
\centering
\begin{minipage}[t]{0.48\textwidth}
\centering
\caption{Contact Localization.}
\label{tab:contact_localization}
\small
\begin{tabular}{lc}
\toprule
Method & RMED $\downarrow$ \\
\midrule
DiffSound~\cite{diffsound} & 41.78\% \\
\textbf{Ours} & \textbf{34.61\%} \\
\bottomrule
\end{tabular}
\end{minipage}
\hfill
\begin{minipage}[t]{0.48\textwidth}
\centering
\caption{Sound Editing.}
\label{tab:sound_editing}
\small
\begin{tabular}{lc}
\toprule
Method & UMAP $\downarrow$ \\
\midrule
Generation & 3.077 \\
Audio-SDS & 4.234 \\
\textbf{Ours} & \textbf{2.753} \\
\bottomrule
\end{tabular}
\end{minipage}
\end{table*}

We demonstrate the utility of \ourmethod through two downstream applications, validating that it captures objects' spatial acoustic properties and supports impact-sound-driven tasks such as contact localization and object sound editing.

\paragraph{Contact Localization.} We evaluate the ability of \ourmethod to infer the 3D impact position $\mathbf{x}^*$ from a novel impact sound recording. We use \textsc{DiffSound}~\cite{diffsound} as our primary baseline, as it also employs a modal-based optimization approach. The mode-specific gains are extracted from the audio and matched against our learned spatial gain field $\mathcal{G}_\theta(\mathbf{x})$ using cosine distance. Notably, real-world contact localization from impact sounds is challenging due to the environmental noise, uniform material property, and non-modal effect introduced during data collection process. It is especially ambiguous for highly symmetric and small objects. Therefore, we report the experiment results on the same subset of non-symmetric objects used in Tab.~\ref{tab:nonsym}. As shown in \cref{tab:contact_localization}, our method substantially outperforms \textsc{DiffSound}, which struggles to localize contacts accurately when its estimated physical parameters are incorrect. We also show heatmap visualizations in \cref{fig:contact_sound_editing_visualization}. Object 1 and 2 exhibit accurate spatial distributions that align well with the ground-truth contact locations, while Object 3 illustrates a failure case caused by high symmetry. See more examples in Supp.

\paragraph{Object Sound Editing.} We evaluate the effectiveness of our Audio-SDS framework for text-driven material editing, with an emphasis on the semantic alignment with the editing prompt. We compare our hierarchical optimization scheme against text-to-audio generation and Audio-SDS~\cite{richter2025audiosds}. As reported in \cref{tab:sound_editing}, our method achieves the lowest UMAP distance, significantly outperforming both the Generation baseline and Audio-SDS (see Supp. for the metric definition and setup). This demonstrates that our hierarchical frequency optimization scheme accurately shifts the acoustic profile to match the target material. In contrast, unconstrained Audio-SDS independently updates individual modal parameters, failing to achieve effective frequency editing across contrastive material pairs. Furthermore, in ~\cref{fig:contact_sound_editing_visualization}, we visualize an editing example---transforming a ceramic object into a contrasting wooden object---by extracting frequency distributions with our modal parameter extractor (\cref{sec:sound_modeling}) and embedding the impact sounds before and after editing into a low-dimensional space using UMAP. We further contextualize this result by overlaying it on the UMAP embedding of 26 objects spanning diverse material properties. We observe that our method can successfully edit the material type of the object. See Supp. for more examples.
\section{Conclusion}
We presented \ourmethod, an audio-visual modal sound representation for object impact sound synthesis. By leveraging visual cues from multi-view image observations and physical parameters estimated from a few impact sound recordings, \ourmethod predicts impact sounds significantly more accurately than prior methods. Beyond impact sound synthesis, our representation also enables new downstream applications involving object impact sounds, including contact localization and object sound editing. Notably, existing real-world object impact sound datasets contain only objects with uniform materials. As a result, we focused on modeling impact sounds under this setting, while leaving generalization to non-uniform materials as an interesting direction for future work.

\newpage
\bibliographystyle{splncs04}
\bibliography{main}

\newpage
\appendix

In this supplementary material, we first expand on our method formulations in \cref{supp:method_details} and detail our implementation specifics in \cref{supp:implementation_details}. Next, we present further experimental analyses in \cref{supp:additional_experiments}. Finally, we provide extended quantitative and qualitative evaluations for our downstream applications in \cref{supp:application}.

\section{Method Details}
\label{supp:method_details}

\subsection{Multi-View Visual Feature Extraction and Refinement}
\label{sec:supp_visual_processing}
To guide acoustic prediction with geometry-aware and appearance semantic guided visual priors, we construct a 3D feature field from multi-view RGB images. Starting from the input images $\mathcal{I}=\{(I_i,\Pi_i)\}_{i=1}^{I}$, we first reconstruct a 3D Gaussian Splatting~\cite{gausssplatting} representation of the object, which yields a dense set of Gaussian centers $
\mathcal{P} = \{\mathbf{o}_i\}_{i=1}^{O}, \mathbf{o}_i \in \mathbb{R}^3.$
These Gaussian centers is further considered as the cloud point which contained the geometry information we needed for reconstruction.

\paragraph{Lifting 2D Visual Features to 3D.}
Besides, the positional information encoded in the cloud point, we would like to provide each point with semantic information from the RGB. Thus, we extract dense 2D visual embeddings from each image using a pretrained DINOv2 encoder~\cite{oquab2024dinov}. Using the known camera parameters $\Pi_i$ and the reconstructed 3DGS geometry, we lift these multi-view features to the Gaussian centers, assigning each center $\mathbf{o}_i$ a feature vector $\mathbf{f}_i \in \mathbb{R}^D$. 

\paragraph{Automatic Symmetry Detection.}
Although the lifted features are geometry-aware, they may still be inconsistent across symmetric parts due to viewpoint variation, occlusion, or appearance ambiguity. For example, on a rotationally symmetric object such as a bowl, points lying on the same rotational orbit should ideally have similar semantic features, yet their lifted features may differ in practice. To enforce intrinsic geometric consistency, we detect object symmetries and refine the lifted 3D features accordingly.

We consider two common symmetry types: rotational symmetry and planar mirror symmetry. A rotational symmetry around axis $\mathbf{a}$ is represented by
$$
\mathcal{T}_{\mathrm{rot}}^{(\theta)}(\mathbf{o})
=
R_{\mathbf{a}}(\theta)\mathbf{o},
\qquad \theta \in [0,2\pi),
$$
where $R_{\mathbf{a}}(\theta)\in SO(3)$ is the rotation matrix around axis $\mathbf{a}$. A mirror symmetry with plane normal $\mathbf{n}$ is represented by
$$
\mathcal{T}_{\mathrm{ref}}(\mathbf{o})
=
\mathbf{o}-2(\mathbf{o}^{\top}\mathbf{n})\mathbf{n}.
$$

\paragraph{Symmetry-aware Feature Refinement.}
The valid symmetries and their axis can be identified by off-the-shelf methods. After determining the symmetries, we refine the 3D features by pooling over symmetric counterparts. For rotational symmetry around axis $\mathbf{a}$, we aggregate features along the sampled rotational orbit:
\begin{equation}
\mathbf{f}_i^{\mathrm{rot}}
=
\frac{1}{|\Theta|}
\sum_{\theta \in \Theta}
\mathrm{Pool}\!\left(\mathcal{T}_{\mathrm{rot}}^{(\theta)}(\mathbf{o}_i)\right),
\end{equation}
where $\Theta$ is a set of sampled rotation angles and $\mathrm{Pool}(\cdot)$ retrieves the feature of the nearest matched Gaussian center after transformation.

For mirror symmetry, we average features between a point and its reflected counterpart:
\begin{equation}
\mathbf{f}_i^{\mathrm{ref}}
=
\frac{1}{2}
\left(
\mathbf{f}_i
+
\mathrm{Pool}\!\left(\mathcal{T}_{\mathrm{ref}}(\mathbf{o}_i)\right)
\right).
\end{equation}

When multiple valid symmetries exist, we refine the visual feature field through an iterative symmetry-consistency procedure. At each refinement step $t$, we first apply the orbit-based refinement and then perform the mirror-based refinement on the updated features. After $T$ refinement steps, the feature associated with point $i$ is denoted as $\mathbf{f}_i^{(T)}$, and the final symmetry-consistent feature is defined as
$
\mathbf{f}_i^{\text{refined}} = \mathbf{f}_i^{(T)}.
$
The resulting refined visual field is
\begin{equation}
\mathcal{V}^{\text{refined}}
=
\{(\mathbf{o}_i,\mathbf{f}_i^{\text{refined}})\}_{i=1}^{O},
\end{equation}
which serves as the visual prior for the downstream audio-visual modal sound field reconstruction.

\subsection{Physically Grounded Object Sound Editing}
\label{sec:physically_grounded_sound_editing}
While Audio-SDS successfully modifies the acoustic parameters $\{f_i, d_i\}_{i=1}^{N}$ and reference gains $\{g_i\}_{i=1}^{N}$ to match a text prompt, we must ensure these edits is consistent with the linear modal analysis to preserve the object's acoustic physics. Building on linear modal analysis, we exploit how material parameters affect the linear modal system. Following similar observations as in~\cite{diffsound}, 
for a fixed geometry and Poisson ratio $\nu$, the Mass and Stiffness matrix satisfy $\mathbf{M}=\rho\mathbf{M}_0$ and $\mathbf{K}=E\mathbf{K}_0$, where $\mathbf{M}_0$ and $\mathbf{K}_0$ depend only on shape and $\nu$. The generalized eigenproblem $\mathbf{K}\mathbf{U}=\mathbf{M}\mathbf{U}\mathbf{S}$ then becomes
\begin{equation}
E\mathbf{K}_0\boldsymbol{\phi}_k=\lambda_k \rho\mathbf{M}_0\boldsymbol{\phi}_k,
\end{equation}
implying that the modal shapes $\boldsymbol{\phi}_k$ remain unchanged, while the eigenvalues scale as $\lambda_k\propto E/\rho$. Under Rayleigh damping, the $k$-th modal coordinate in 
\begin{equation}
    s(t)=\sum_{i=1}^{N} g_i e^{-d_i t}\sin(2\pi f_i t),
    \label{eq:sinusoids}
\end{equation}
is an independent damped oscillator with decay rate $\delta_k=\frac{1}{2}(\alpha+\beta\lambda_k)$ and damped frequency $f_k=\sqrt{\lambda_k-\delta_k^2}/2\pi$. Therefore, editing $(E,\rho,\alpha,\beta)$ changes the modal frequencies and dampings, but not the spatial mode shapes. For an impact at location $x$ with direction $\mathbf d$, the modal gain in Eq.~\eqref{eq:sinusoids} can be written as $a_k(x) = \frac{\boldsymbol{\phi}_k(x)^\top \mathbf d}{2\pi f_k};$
so after editing material parameters, the gain updates via a simple per-mode rescaling:
\begin{equation}
a_k^{(2)}(x) = a_k^{(1)}(x)\frac{f_k^{(1)}}{f_k^{(2)}}.
\end{equation}
Moreover, for two impact locations $x$ and $y$ under the same direction $\mathbf d$, the gain ratio $\frac{a_k(x)}{a_k(y)}=\frac{\boldsymbol{\phi}_k(x)^\top \mathbf d_x}{\boldsymbol{\phi}_k(y)^\top \mathbf d_y}$
is invariant to such parameter editing, since the frequency-dependent factor cancels out. This property allows us to preserve the learned spatial modal field and perform physically grounded sound editing by only updating the per-mode frequencies, decay rates, and gain scales.

\begin{figure}[t!]
    \centering
    \includegraphics[width=0.8\linewidth]{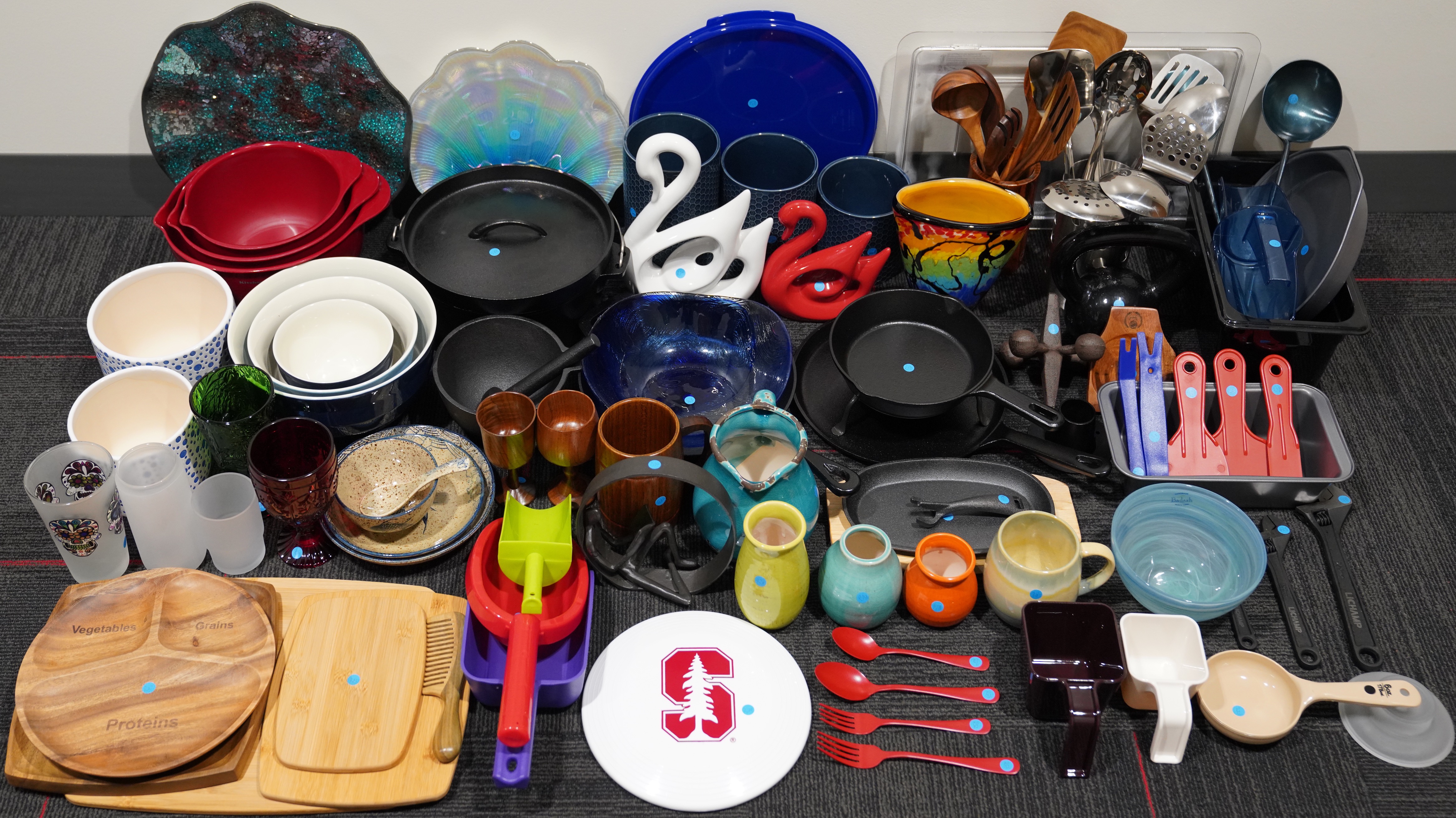}
    \caption{Illustration of Objects in the \textsc{ObjectFolder Real} Dataset~\cite{gao2023objectfolder}.}
    \label{fig:real_objects}
\end{figure}

\section{Implementation Details}
\label{supp:implementation_details}
\paragraph{Training Setup Details.}
For each object, we follow the same reconstruction pipeline as described in SonicGauss to obtain a 3D Gaussian Splatting representation. We then extract dense per-image visual features using DINOv2 and lift them into 3D using the known camera poses. Specifically, each Gaussian center is projected into all visible training views. For each valid projection, we sample the corresponding DINOv2 feature map using bilinear interpolation, and average the sampled descriptors across views to obtain a per-Gaussian embedding. We then refine these lifted features using symmetry-aware pooling, the final feature field are computed on the full set of Gaussian centers.

For audio clustering, the number of clusters is selected from \{128, 256, 512\} for each object to balance efficiency and performance. We optimize the model with Adam using a learning rate selected from \{1e-2, 5e-3, 1e-3\} and train for a number of epochs selected from \{5000, 10000, 20000\}. A cosine learning-rate scheduler is used with a minimum learning rate of 0.1 times the initial learning rate. The cutoff loss weight is selected from \{0.5, 1, 2\}. For each object, we run both the single-damping and spatial-damping variants, and choose the best model based on the ENV metric. We use multi-scale STFT losses with FFT sizes 2048, 1024, 512, and 256, and a hop size of 64. The training can be conducted with a single NVIDIA A5000 GPU.

\paragraph{Ablation Subset Details.}
For the ablation study, we use 10 objects from ObjectFolder Real, with object IDs 1, 11, 28, 38, 41, 44, 52, 57, 66, and 75. This subset covers a diverse range of materials and shapes, including objects that produce both low- and high-frequency impact sounds, as well as both symmetric objects (e.g., a cup) and non-symmetric objects (e.g., a spoon). We also use this subset for \cref{supp:sampling_strategies}, \cref{supp:few_shot}, and \cref{supp:input_views}.

\paragraph{Non-symmetric Objects Subset Details.}
As shown in \cref{fig:real_objects}, the ObjectFolder Real dataset is mainly composed of symmetric objects. To validate our method’s ability to learn spatial modal parameters, we select a subset of 16 non-symmetric objects from ObjectFolder Real, with IDs 1, 14, 17, 24, 39, 40, 41, 44, 45, 52, 54, 55, 57, 66, 79, and 91. We use these subset for a case study and quantitative evaluation of contact localization.

\begin{figure}[t!]
\centering
\includegraphics[width=\linewidth]{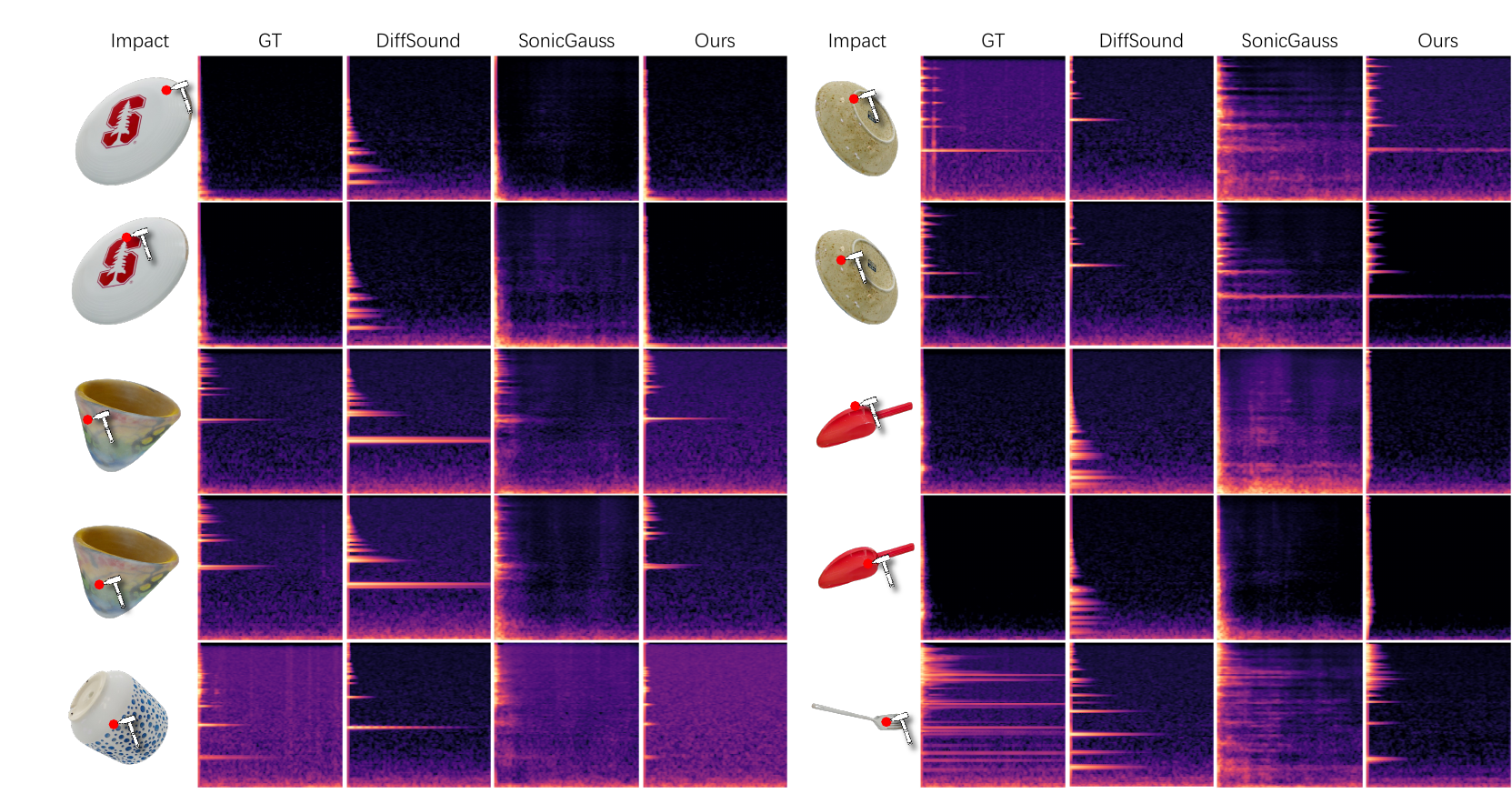}
\caption{\textbf{Spectrogram Comparisons.} We provide additional spectrogram comparisons between the baselines and our method. All spectrograms are generated using the same settings as in the main paper. In most cases, our method yields more accurate and sharper reconstructions. Two representative failure cases are included at the bottom for completeness.} 
\label{fig:more_spectrogram_compare}
\end{figure}

\section{Additional Experiments}
\label{supp:additional_experiments}
\subsection{Computation.}
On a single A5000 GPU, our method trains in $0.27$h and supports real-time inference at $43.8$ms. In comparison, SonicGauss takes $0.59$h per object and $3.2$s for inference, while DiffSound takes $3.16$h for training and $6.9$ms for inference. 

\subsection{Qualitative Spectrogram Comparisons.}
We includes more qualitative comparison results in Fig.~\ref{fig:more_spectrogram_compare}. In most cases, our method provides more accurate and sharper results than the baselines. We also include two failure cases at the bottom row of the figure. For both cases, the queried impact locations lie near regions with sharp geometric changes, yet no training samples are collected from similar local geometries due to the farthest-point sampling strategy. Thus performance is limited by the lack of representative training observations, leading to short damping. In contrast, SonicGauss~\cite{sonicgauss} can better handle these cases by leveraging knowledge acquired from large-scale pretraining.

\begin{table}[t!]
\centering
\caption{Case Study Results on Sampling Strategies for Selecting Training Locations.}
\label{tab:sample_strategies}
\begin{tabular}{lcccc}
\toprule
Sampling Strategy & L1 & L1 Log & ENV & CDPAM \\
\midrule
FPS & 0.019 & 0.927 & 0.015 & 2.00e-4 \\
CPS & 0.019 & 1.010 & 0.015 & 2.08e-4 \\
\bottomrule
\end{tabular}
\end{table}

\subsection{Case Study on Sampling Strategies}
\label{supp:sampling_strategies}
We use farthest point sampling (FPS) as the default strategy for selecting the training set, which simulates uniformly collecting impact samples over the object surface. To further evaluate the transfer ability of our method, we additionally consider closest point sampling (CPS), where training samples are concentrated within a limited spatial region. As shown in Tab.~\ref{tab:sample_strategies}, under this more challenging setting, the performance of ours remain stable. This suggests that our method benefits from the visual prior and generalizes better beyond the observed training area.

\begin{table}[t!]
\centering
\caption{Comparing N-shot performance, L1 peformance is reported.}
\label{tab:fewshot}
\begin{tabular}{lccc}
\toprule
Shot Number 
&DiffSound 
&SonicGauss
&Ours \\
\midrule
16 & 0.0254 & 0.0245 & 0.0143 \\
8 & 0.0253 & 0.0259 & 0.0153 \\
4 & 0.0253 & 0.0336 & 0.0224 \\
2 & 0.0255 & 0.0855 & 0.0223 \\
1 &  0.0254 & 0.0952 & 0.0256 \\
\bottomrule
\end{tabular}
\end{table}

\subsection{Few-shot Experiments}
\label{supp:few_shot}
A key advantage of physics-based models is their ability to learn effectively in few-shot, or even one-shot, settings, since the underlying parameters can in principle be estimated from only a small number of audio recordings. We show that AV-MSF inherits this advantage. As shown in Tab.~\ref{tab:fewshot}, performance degrades only slightly when the number of shots decreases from 16 to 8, while drops more noticeably from 8 to 4, and remains stable from 4 to 1. This suggests that 4--8 shots already provide sufficient signal to capture the main acoustic profile. Moreover, the strong performance under 1--2 shots demonstrates the benefit of leveraging visual priors in our method. In contrast, the data-driven baseline performs poorly in the low-shot regime. The physics-based baseline remains relatively stable, but due to the limitations of simulation-based rendering, its performance is still worse than ours in few-shot setting, and comparable with one-shot. Although the diffusion-based model improves as more training data becomes available, its performance remains worse than ours, likely due to its nature to generate hallucinate unrealistic signals.

\subsection{Comparison across different numbers of input views}
\label{supp:input_views}
We compare performance when using fewer images to reconstruct the 3D Gaussian prior. Reducing the number of input views leads to a noisier point cloud and less informative visual features, which weakens both the geometric and appearance priors. As shown in \cref{tab:input_views}, the overall performance degrades, highlighting the importance of high-quality geometric and visual priors in our framework. Nevertheless, the performance drop remains relatively modest, which demonstrates the robustness of our method even under reduced multi-view supervision.

\begin{table}[t!]
\centering
\caption{\textbf{Effect of input view numbers.} We compare different number of input views used for 3D prior reconstruction. More views can provide better vision prior and give better performance.}
\label{tab:input_views}
\begin{tabular}{lcccc}
\toprule
Image Number & L1 & L1 Log & ENV & CDPAM \\
\midrule
12 & 0.020 & 1.002 & 0.016 & 2.84e-4 \\
71 (full) & \textbf{0.019} & \textbf{0.927} & \textbf{0.015} & \textbf{2.00e-4} \\
\bottomrule
\end{tabular}
\end{table}

\section{Additional Application Results}
\label{supp:application}

\subsection{Contact Localization}
\label{supp:contact_localization}

\begin{figure}[t!]
\centering
\includegraphics[width=\linewidth]{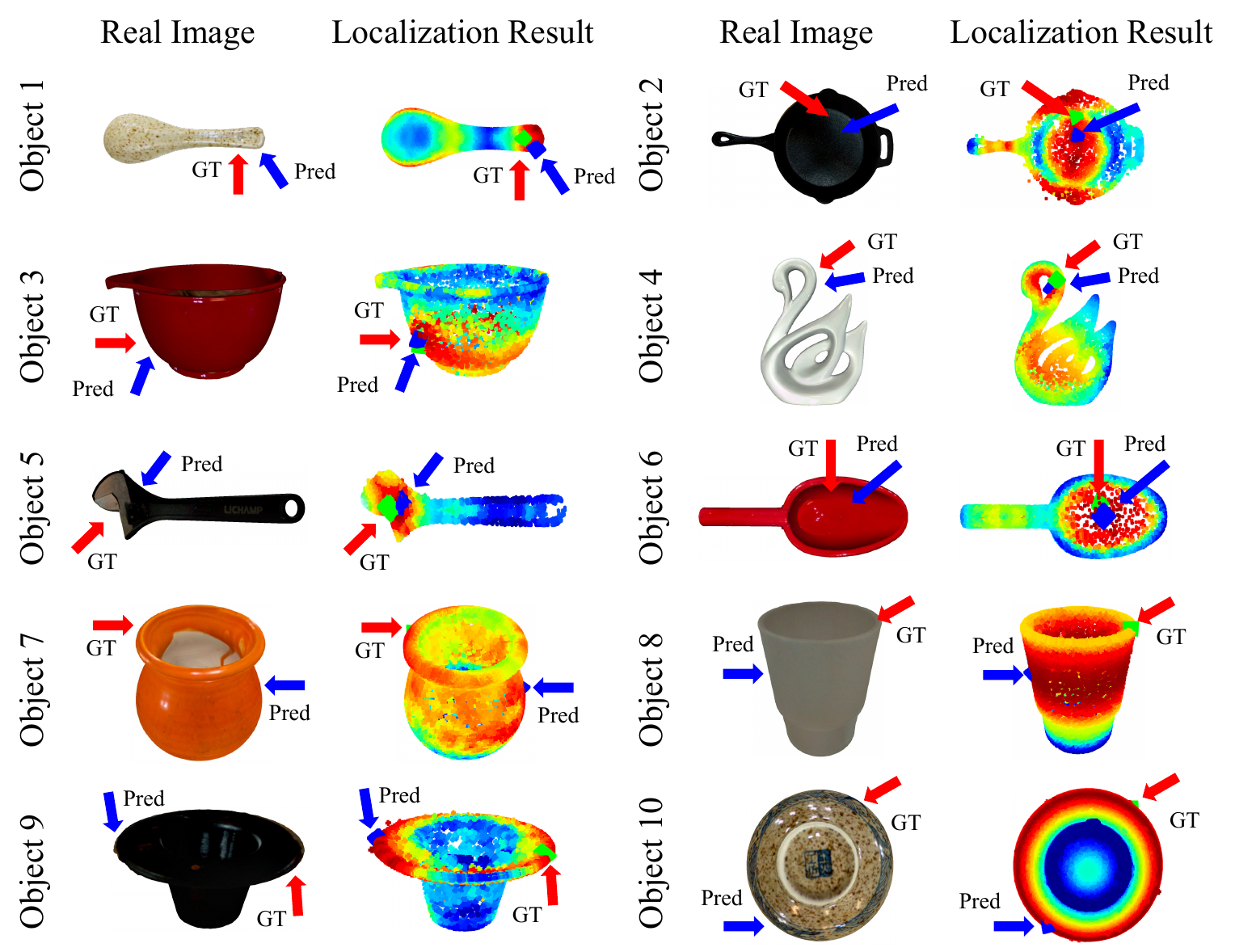}
\caption{\textbf{Contact Localization Heatmaps.} Predicted regions are shown with red indicating high likelihood and blue indicating low likelihood. Objects 7, 8, 9, and 10 show failure cases.} 
\label{fig:contact_compare}
\end{figure}

As discussed in the main text, we evaluate our method's ability to infer 3D impact locations from novel sound recordings against \textsc{DiffSound}~\cite{diffsound}.

\paragraph{Implementation Details.}
We use the same data subset of non-symmetric study (\cref{supp:implementation_details}). For \textsc{DiffSound}, we use the original ObjectFolder Real mesh to simulate the gain at every vertex.

\paragraph{Qualitative Results.}
\cref{fig:contact_compare} provides additional visualizations across diverse objects, including both successful and failure cases. The heatmaps represent the cosine distance between gains extracted from the input audio and our reconstructed spatial gain field. For less symmetric objects (Objects 1, 4, 5, and 6) and larger objects (Objects 2 and 3), impact sounds exhibit more diverse patterns that better correlate with geometry. However, contact localization from object impact sounds remains challenging for highly symmetric or small objects. For symmetric objects (Objects 9 and 10), symmetric regions such as edges can exhibit similar modal parameters while being spatially far apart. For small objects (Objects 7 and 8), the vibration modes are often too simple, causing impacts at different locations to produce similar sounds. Consequently, the heatmaps show high likelihood across the whole object.

\subsection{Object Sound Editing}
\label{supp:sound_editing}
This section provides further evaluations of our object sound editing approach. We compare our method against two baselines: direct text-to-audio \textbf{Generation} (generating sound directly from a text prompt via a pretrained diffusion model) and unconstrained \textbf{Audio-SDS}~\cite{richter2025audiosds}, which independently optimizes the modal parameters $\{f_i\}_{i=1}^N$.

\paragraph{Evaluation Metrics.}
We assess editing quality using two primary metrics:
\begin{itemize}
\item \textbf{CLAP Score:} Measures the semantic alignment between the edited impact sound and the target text prompt (higher is better).
\item \textbf{UMAP Distance:} Quantifies how accurately the edited frequencies reflect the target material type. We compute the average distance in UMAP space between the edited frequencies and the target material's cluster center across various material mappings and prompts (lower is better).
\end{itemize}

\paragraph{Implementation Details.}
We follow the setup of \cite{richter2025audiosds}, with two key modifications. First, instead of using l2 loss, we employ a multi-scale STFT loss to better capture frequency-domain characteristics. Second, unlike Audio-SDS, which defaults to optimizing 2048 modes, we initialize our modal parameters directly from our pretrained modal sound field. Using 2048 modes is computationally prohibitive in our pipeline due to the overhead introduced by the attention mechanism. Instead, we use 128 modes, which provides sufficient granularity for high-fidelity impact sound synthesis.

\begin{figure}[t!]
\centering
\includegraphics[width=\linewidth, height=0.9\textheight, keepaspectratio]{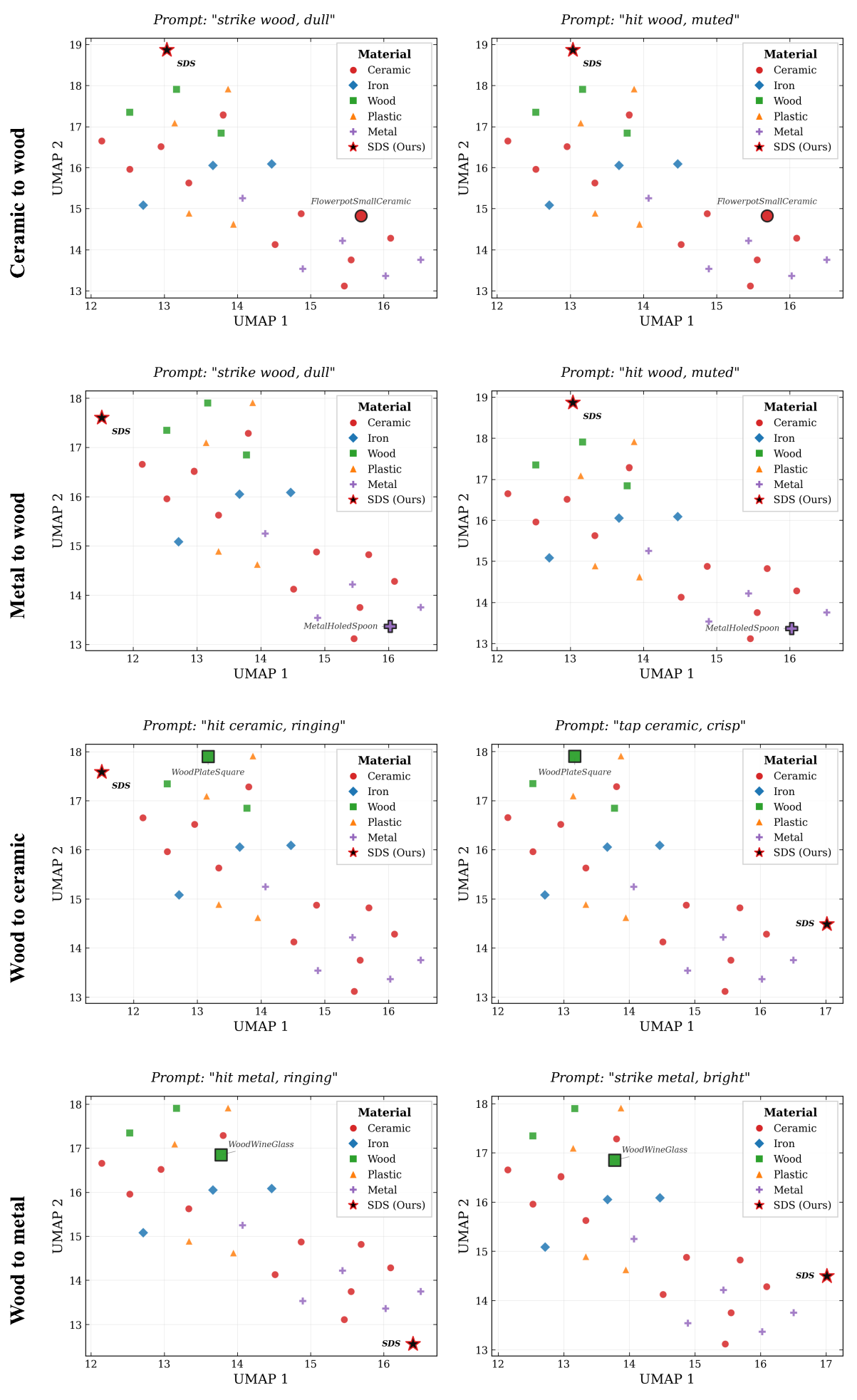}
\caption{\textbf{Sound Editing in UMAP Space.} Modal frequencies consistently shift toward target material clusters post-editing.}
\label{fig:sound_editing_umap}
\end{figure}

For evaluation, we select 26 objects with material labels from the RealImpact dataset. We extract their modal frequencies using our modal parameter extractor (Sec.~3.4 in main) to serve as ground-truth points in the UMAP space. To evaluate the robustness of our method, we test edits across contrastive material transitions with two text prompts. Specific details are provided in \cref{fig:sound_editing_umap}. 

For the Generation baseline in the UMAP distance comparison, we similarly apply the modal parameter extractor described in Sec.~3.4 in main to obtain modal frequencies from the directly generated sounds.

\begin{figure}[t!]
    \centering
    \includegraphics[width=0.8\linewidth]{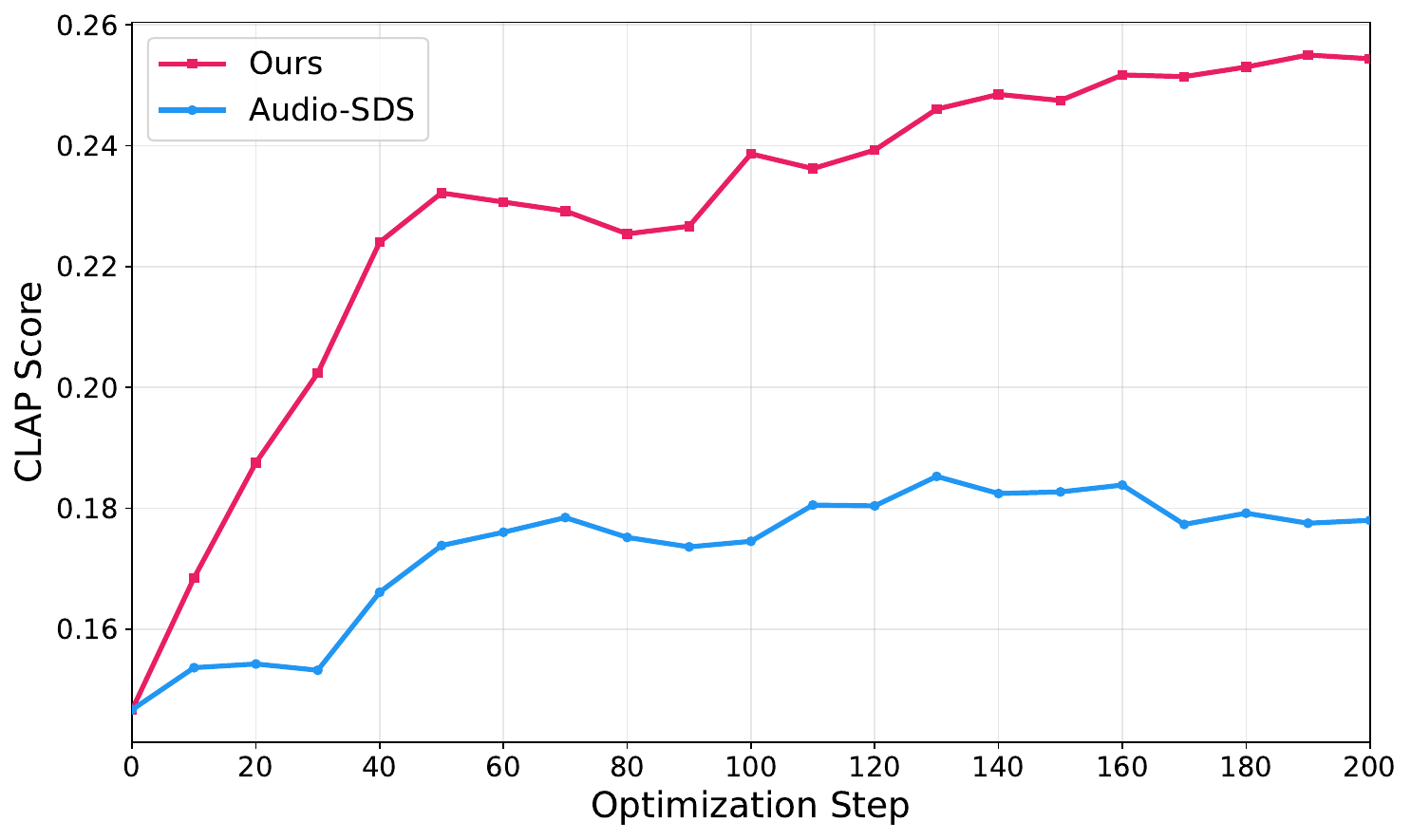}
    \caption{\textbf{CLAP Score Progression.} Semantic alignment consistently improves during editing.}
    \label{fig:clap_compare}
\end{figure}

\paragraph{Quantitative Results.} 
\cref{fig:clap_compare} illustrates that our approach yields consistent improvements in CLAP score during training, indicating that the multi-scale STFT loss and our hierarchical frequency optimization successfully update the modal parameters toward the target semantics.

\paragraph{Qualitative Results.} 
\cref{fig:sound_editing_umap} visualizes the extracted modal frequencies within the UMAP space before and after editing. Our results consistently shift toward the target material clusters, demonstrating effective sound editing across contrastive pairs. Notably, some ground-truth ceramic points fall closer to the wood cluster; this is primarily because some ceramic objects has solid shape, which naturally exhibit lower-frequency impact sounds. However, we also observe a failure case in the "wood to ceramic" editing, where the resulting frequencies fall far from the ceramic cluster, highlighting the inherent variance of generative-based methods.

\end{document}